\documentclass[10pt,conference,letterpaper]{IEEEtran}
\IEEEoverridecommandlockouts

\usepackage[T1]{fontenc}
\usepackage{amsmath,amssymb}
\usepackage{graphicx}
\graphicspath{{./}{demo/}{figures/}}
\usepackage{diagbox}
\usepackage{float}
\usepackage{xcolor}
\usepackage{booktabs}
\usepackage{cite}
\usepackage{url}
\usepackage{balance}

\floatstyle{ruled}
\newfloat{algorithm}{t}{loa}
\floatname{algorithm}{Algorithm}
\newcounter{alglnum}
\newenvironment{pcode}{%
  \begin{list}{{\scriptsize\bfseries\arabic{alglnum}:}}{%
     \usecounter{alglnum}%
     \setlength{\leftmargin}{2.0em}\setlength{\labelsep}{0.6em}%
     \setlength{\labelwidth}{1.4em}\setlength{\itemsep}{1.2pt}%
     \setlength{\parsep}{0pt}\setlength{\topsep}{4pt}}%
}{\end{list}}
\newcommand{\ind}[1]{\hspace*{#1em}}
\newcommand{\kw}[1]{\textbf{#1}}
\newcommand{\arw}{\allowbreak\ensuremath{\rightarrow}\allowbreak}

\begin{document}

\title{Enhancing Deep Reinforcement Learning for\\Robotic Motion Planning via Task Decomposition%
\thanks{The paper is accepted for IEEE MSN 2026 conference.}\thanks{Code: \protect\url{https://github.com/YuanzhiHe/panda_gym_obstacle}}}

\author{\IEEEauthorblockN{Yuanzhi He, Victor Romero-Cano, Jiayu Xu, Juan David Hern\'andez, Gualtiero Colombo}
\IEEEauthorblockA{School of Computational and Mathematical Sciences\\
Cardiff University, United Kingdom}}

\maketitle

\begin{abstract}
Obstacle-aware robotic manipulation can require substantial interaction data when deep reinforcement learning policies are trained from scratch. We study a task decomposition framework combining first-layer critic transfer with staged reward redesign (RR) on a Truncated Quantile Critics (TQC) backbone. Transfer reuses an input representation across tasks, while RR supplies intermediate feedback for Pick-and-Place. Experiments cover Push, Pick-and-Place, and Slide in obstacle-free and single-obstacle PyBullet environments, with three random seeds per setting. On obstacle-aware Push, transfer achieves $0.83\pm0.05$ success against $0.77\pm0.06$ for TQC. On obstacle-aware Slide, it achieves $0.95\pm0.02$ against $0.84\pm0.06$ for SAC, the strongest baseline in that setting. These are gains of $6$ and $11$ percentage points, respectively. RR reduces the target-training steps needed for obstacle-free Pick-and-Place from roughly $0.75$M to $0.20$M. Transfer benefits depend on the route: obstacle-aware Pick-and-Place shows a lower mean than training from scratch. A selected simulated trajectory is executed open-loop on a Franka Emika FR3 as a feasibility demonstration. The study characterizes when a lightweight combination of established transfer and reward-design components helps learning.

\end{abstract}


\begin{figure*}[t]
    \centering
    \includegraphics[width=0.19\textwidth]{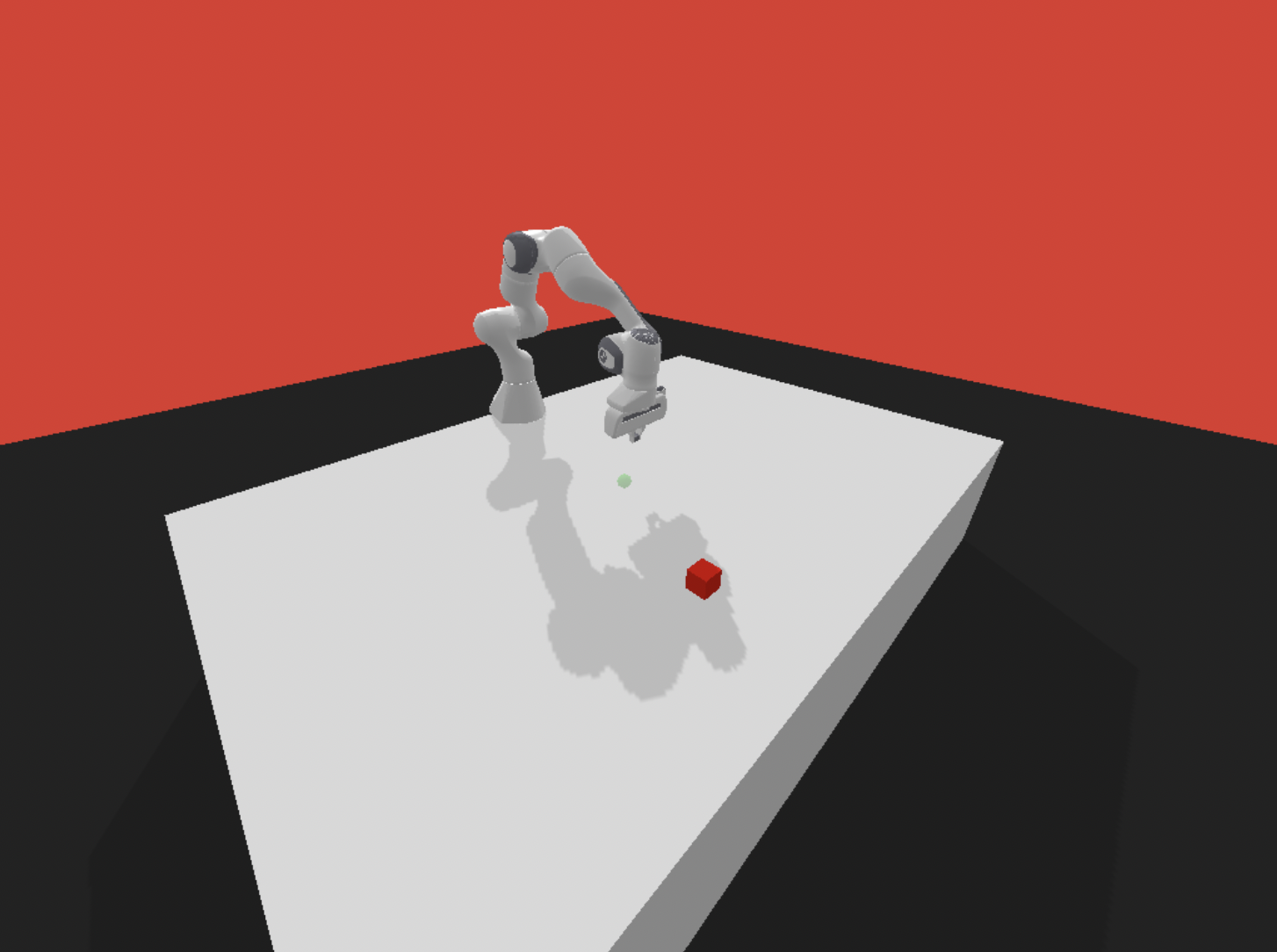}
    \includegraphics[width=0.19\textwidth]{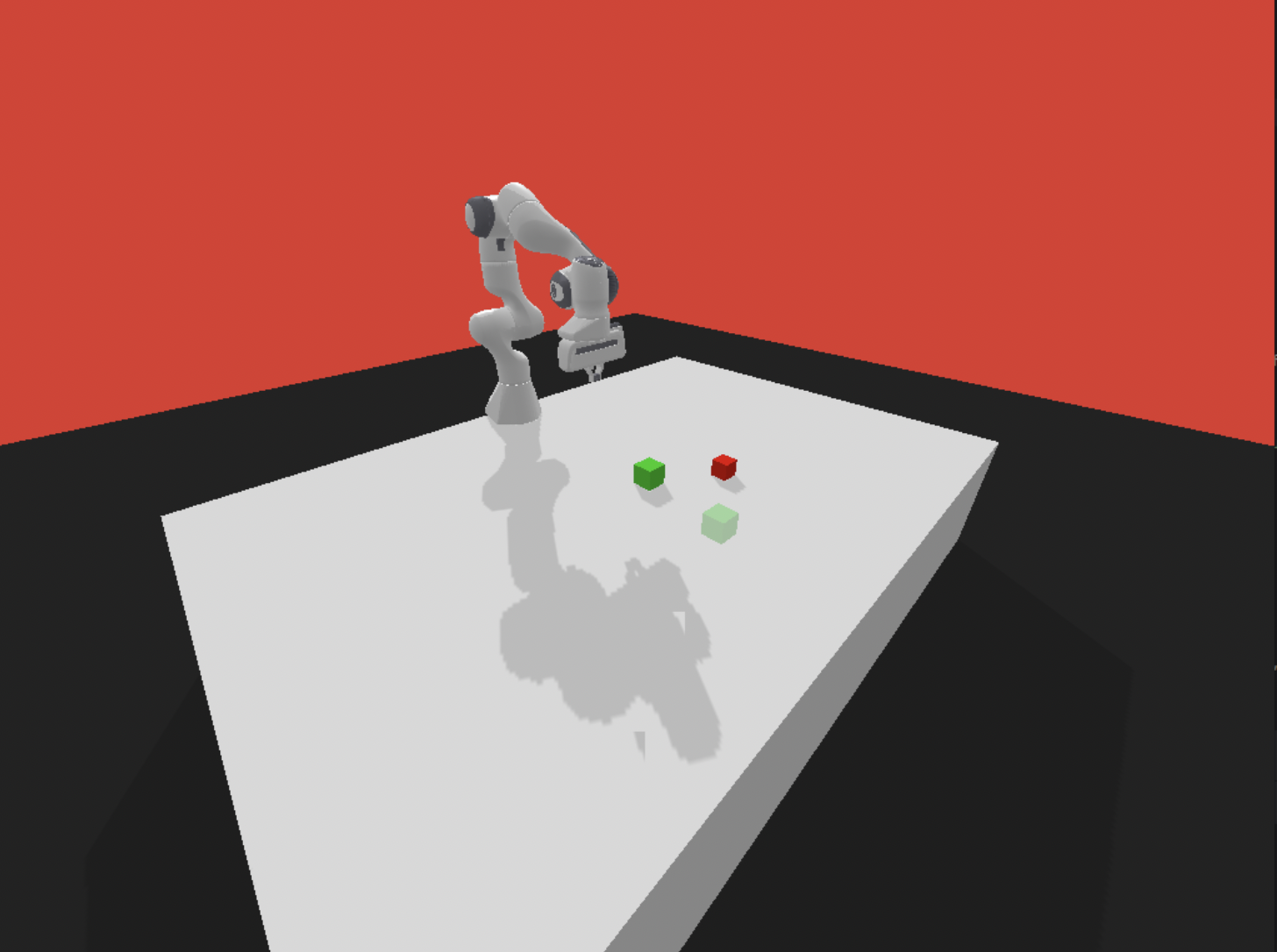}
    \includegraphics[width=0.19\textwidth]{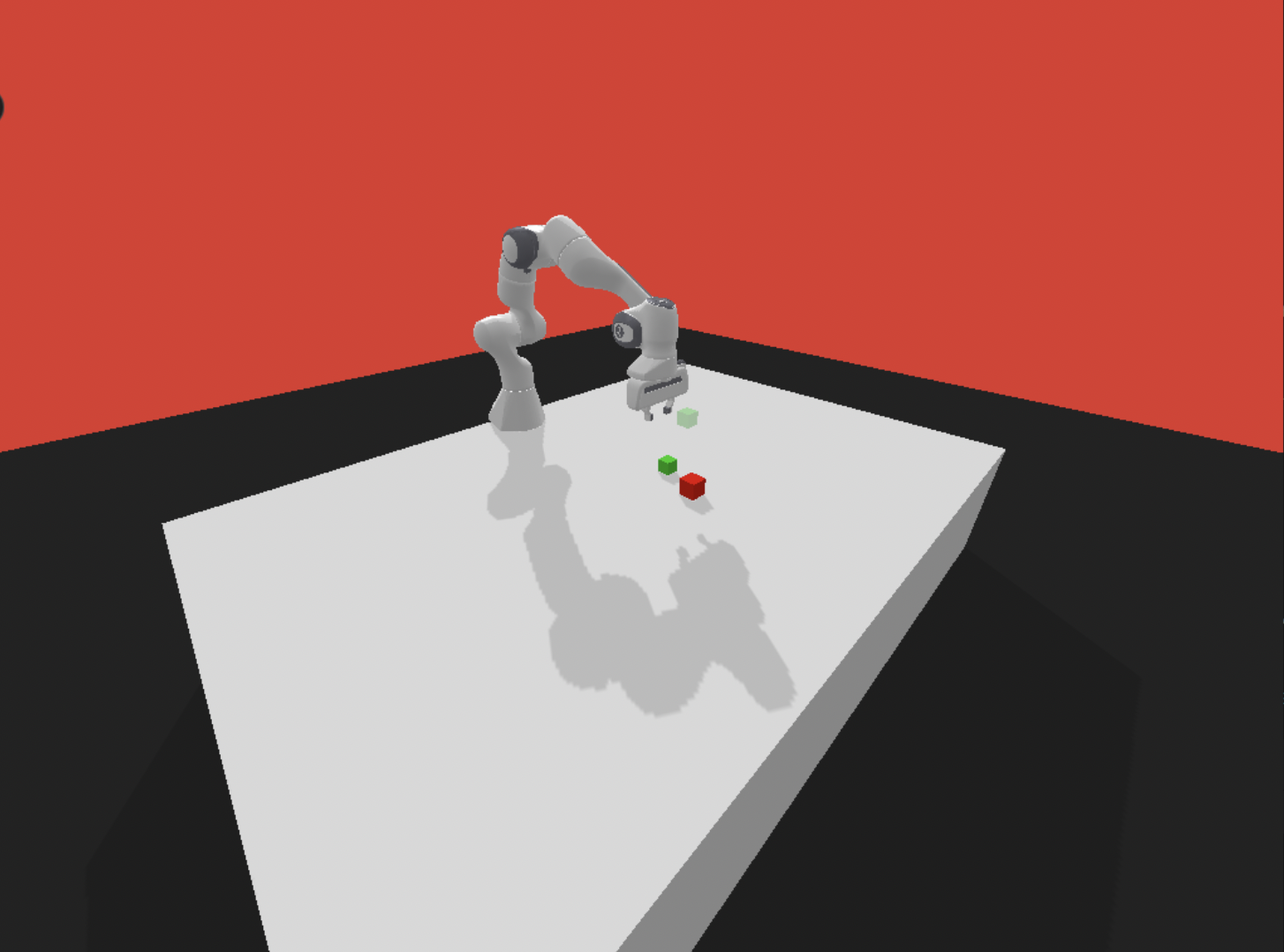}
    \includegraphics[width=0.19\textwidth]{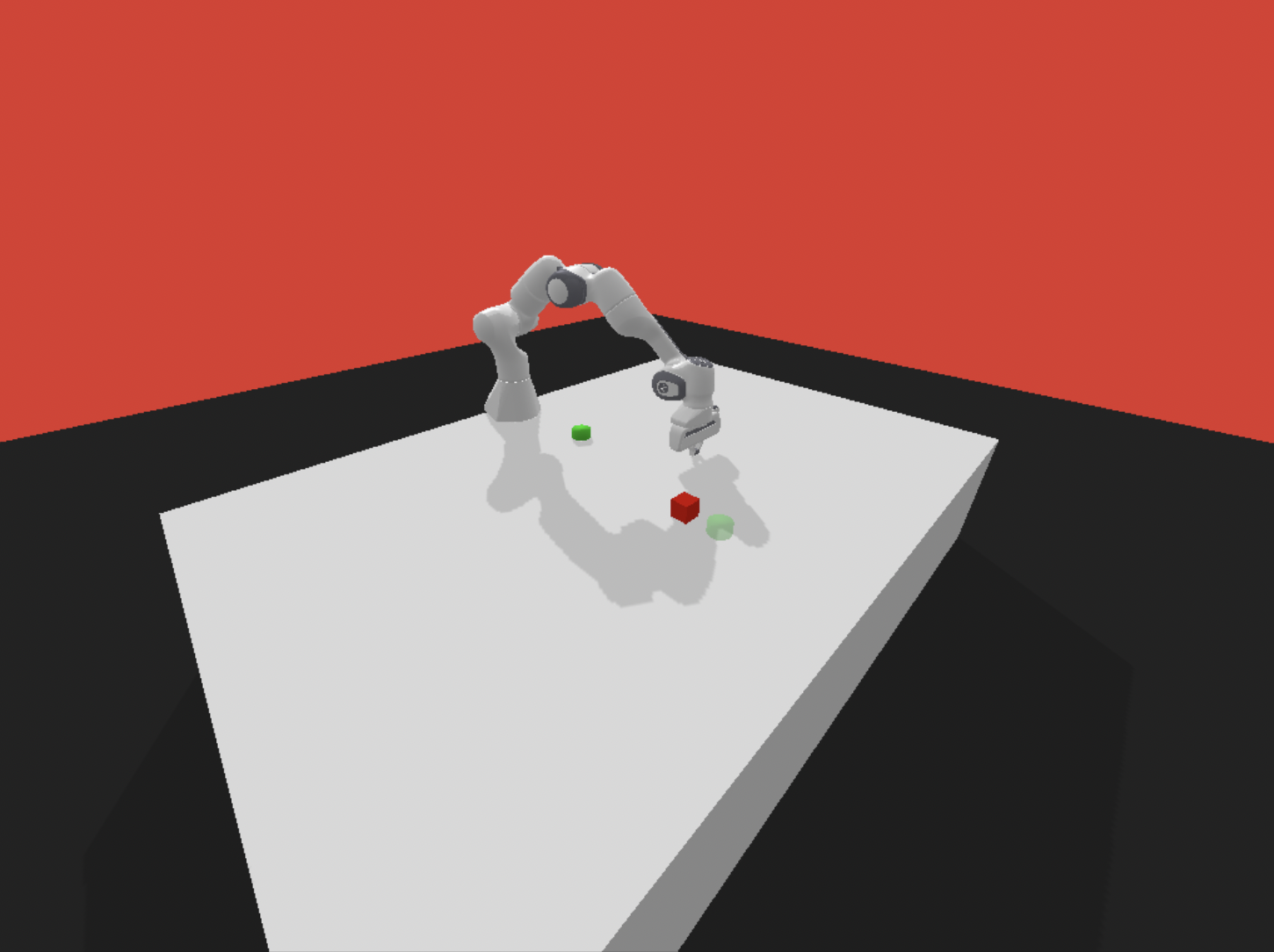}
    \caption{Reach (1st), Push (2nd), Pick\&Place (3rd) and Slide (4th) in PyBullet Panda simulation, each with one red cube obstacle.}
    \label{panda_demos}
    \vspace{-0.2cm}
\end{figure*}

\section{Introduction}

Robotics research has increasingly shifted toward simulation environments rather than training directly on real robots~\cite{tang2025deep}. This trend is mainly driven by the high cost of trial-and-error in real-world settings, safety risks to physical hardware, and the time-consuming and labor-intensive nature of collecting large-scale interaction data. In contrast, simulation platforms enable low-cost experimentation with highly controllable scenarios, making them well suited for large-scale and repetitive training processes~\cite{Todorov2012, Erez2015}.
In this study, we adopt the PyBullet Panda environments, whose robot arm model closely matches the kinematic structure and joint constraints of real manipulators such as the Franka arm~\cite{Yang2021}, building on the goal-conditioned robotics benchmarks established in prior work~\cite{Brockman2016, Plappert2018}.

Despite the success of Deep Reinforcement Learning (DRL) in robotic manipulation tasks such as grasping, pushing, and object transportation, increasing task complexity often leads to a rapid growth in training cost and model instability. In particular, the solution of tasks that require the robot to jointly solve manipulation and collision-free path planning problems enlarges the observation and action spaces. Under such conditions, training policies from scratch typically demands a large number of interaction samples, substantial computational resources, and often suffers from slow or unstable convergence~\cite{Wang2024, Song2022, Han2023}. We address this as obstacle-aware, goal-conditioned manipulation policy learning, in which the policy must reach a goal, interact with the object, and avoid the collision region introduced by an obstacle, all learned implicitly from interaction rather than from an explicit geometric planner.

To address these challenges, we propose a task decomposition framework that integrates Transfer Learning (TL) with Reward Redesign (RR) to improve learning efficiency and performance in a set of obstacle tasks. The policy is first trained on a relatively simpler source task to acquire fundamental knowledge. The learned knowledge is then transferred to a more complex target task by reusing selected network parameters. In pick-and-place environments, we further redesign multi-step task rewards by decomposing them into multiple staged feedback signals corresponding to intermediate sub-tasks, where rewards are assigned to the approaching, grasping, and transporting stages.

We evaluate the proposed framework on three manipulation tasks: Push, Pick-and-Place, and Slide, each with obstacle-free and obstacle settings, where a cubic obstacle was introduced in the obstacle environments. Snapshots of obstacle environments are illustrated in Fig.~\ref{panda_demos}. Through systematic comparisons, we show that few-shot TL together with RR improves success rates and reduces training time across most task settings, with one exception, Pick-and-Place with obstacle, which we discuss in detail.

Although first-layer transfer, reward shaping, and task decomposition build on established ideas, our emphasis is on a minimal and controlled combination of them and on understanding when such transfer helps and when it does not. The main contributions of this paper are summarized as follows:
\begin{itemize}
    \item We propose a unified task decomposition framework for DRL that combines few-shot TL and RR to enhance learning efficiency and performance for obstacle-aware robotic manipulation tasks.
    \item We introduce a controlled transfer strategy that reuses only the first-layer critic parameters across tasks, and compare this initialization against training from scratch across different task and obstacle settings.
    \item We design a staged reward formulation that decomposes multi-step task objectives into interpretable sub-task rewards, accelerating convergence without compromising final task performance.
    \item We conduct extensive evaluations on multiple manipulation tasks in both obstacle-free and obstacle environments, including a systematic study of forward, double, reverse, and neutral transfer, and demonstrate real-world feasibility through a deployment on a Franka Emika FR3 robot.
\end{itemize}

The remainder of this paper is organized as follows. Section~2 reviews related work. Section~3 details the proposed approach. Section~4 presents and discusses the experimental results in both simulation and real-world environments. Section~5 concludes the paper and discusses potential future work.

\section{Related Works}

TL aims to reuse knowledge acquired from a source task to facilitate learning in a related target task, thereby reducing training cost and improving performance~\cite{pan2009survey, Jaquier2025, Zhu2023}. In robotic manipulation, TL is particularly valuable due to the high sample complexity of DRL and the substantial interaction cost required to learn complex contact-rich behaviors.
Specifically, few-shot TL enables partial fine-tuning while preserving reusable representations learned in upstream tasks~\cite{Song2022}. In robotic manipulation, several studies have explored parameter reuse and representation transfer to improve learning efficiency~\cite{Jang2022, he2024, he2026sample}, drawing on multi-task representation learning~\cite{Ruder2017}. Few-shot and zero-shot transfer strategies based on selective parameter sharing have demonstrated improved convergence in manipulation tasks by reusing low-level state--action representations~\cite{he2026sample}. However, most existing approaches primarily focus on obstacle-free or structurally similar environments, and less attention has been given to understanding how transfer behaves under geometric constraints or progressively increasing task complexity~\cite{Jaquier2025, Han2023, Kirk2023}.

In recent years, DRL has emerged as a mainstream approach for robotic manipulation tasks~\cite{Wang2024, Han2023}. DRL algorithms are generally categorized into on-policy and off-policy methods~\cite{Schulman2017}.
On-policy algorithms, such as Proximal Policy Optimization (PPO)~\cite{Schulman2017}, are known for their robustness in policy optimization, although they tend to be less sample-efficient. In contrast, off-policy algorithms, including Deep Deterministic Policy Gradient (DDPG)~\cite{lillicrap2020ddpg}, Twin Delayed Deep Deterministic Policy Gradient (TD3)~\cite{fujimoto2018td3}, Soft Actor-Critic (SAC)~\cite{Haarnoja2018sac}, and Truncated Quantile Critics (TQC)~\cite{Kuznetsov2020tqc}, offer greater sample efficiency by leveraging experience replay and reusing previously collected transitions. This property makes them particularly suitable for large-scale tasks that require prolonged interaction with the environment. To address the challenges posed by sparse rewards and large action spaces, \cite{Andrychowicz2017}~has introduced Hindsight Experience Replay (HER), which replaces the originally intended goal with the goal actually achieved in a given episode before storing it in the replay buffer. This technique significantly improves learning efficiency in sparse-reward settings. We build on TQC as our backbone, as it is the strongest baseline in our experiments.

Foundation models such as CLIP~\cite{radford2021clip}, and DINO~\cite{caron2021dino}, pre-trained on large-scale visual or multi-modal datasets, are increasingly applied in robotics primarily as perception backbones and high-level task representations, rather than as low-level control policies.
Recent work also uses vision-language models as task planners and grounds object-centric motion from human demonstration videos~\cite{huang2024, yin2025object}.
Their main advantage lies in generalizability and sample efficiency, enabling robots to adapt to new objects and environments; however, domain shift between pre-training data and real-world conditions can degrade performance, and computational overhead may limit real-time deployment.

Our framework targets a common bottleneck in this setting: training a DRL policy from scratch for complex, obstacle-rich tasks requires substantial interaction data and long training time. The transfer-based approach we adopt improves sample efficiency through structured knowledge reuse.

\section{Task Decomposition for Deep Reinforcement Learning}

\subsection{Problem Formulation}

We consider a family of robotic manipulation tasks executed by a 7-DoF Franka Panda robotic arm in both obstacle-free and obstacle environments. Each task is formulated as a goal-conditioned Markov Decision Process (MDP)~\cite{Colas2022} defined by
\[
\mathcal{M} = \langle \mathcal{S}, \mathcal{A}, \mathcal{G}, P, R, \gamma \rangle,
\]
where $\mathcal{S}$ denotes the continuous state space, $\mathcal{A}$ is the continuous action space, $\mathcal{G}$ represents the goal space, $P(s_{t+1}\mid s_t,a_t)$ is the transition dynamics, $R(s_t,a_t,g)$ is the reward function conditioned on the goal $g \in \mathcal{G}$, and $\gamma \in (0,1]$ is the discount factor.

The objective of DRL is to learn a goal-conditioned policy $\pi_\phi(a_t \mid s_t, g)$ that maximizes the expected cumulative return:
\begin{equation}
\max_{\pi_\phi}
\mathbb{E}_{\pi_\phi}
\left[
\sum_{t=0}^{T} \gamma^t R(s_t, a_t, g)
\right].
\end{equation}

\subsection{Augmenting TQC with Transfer Learning}

\begin{figure*}[t]
    \centering
    \includegraphics[width=0.70\textwidth]{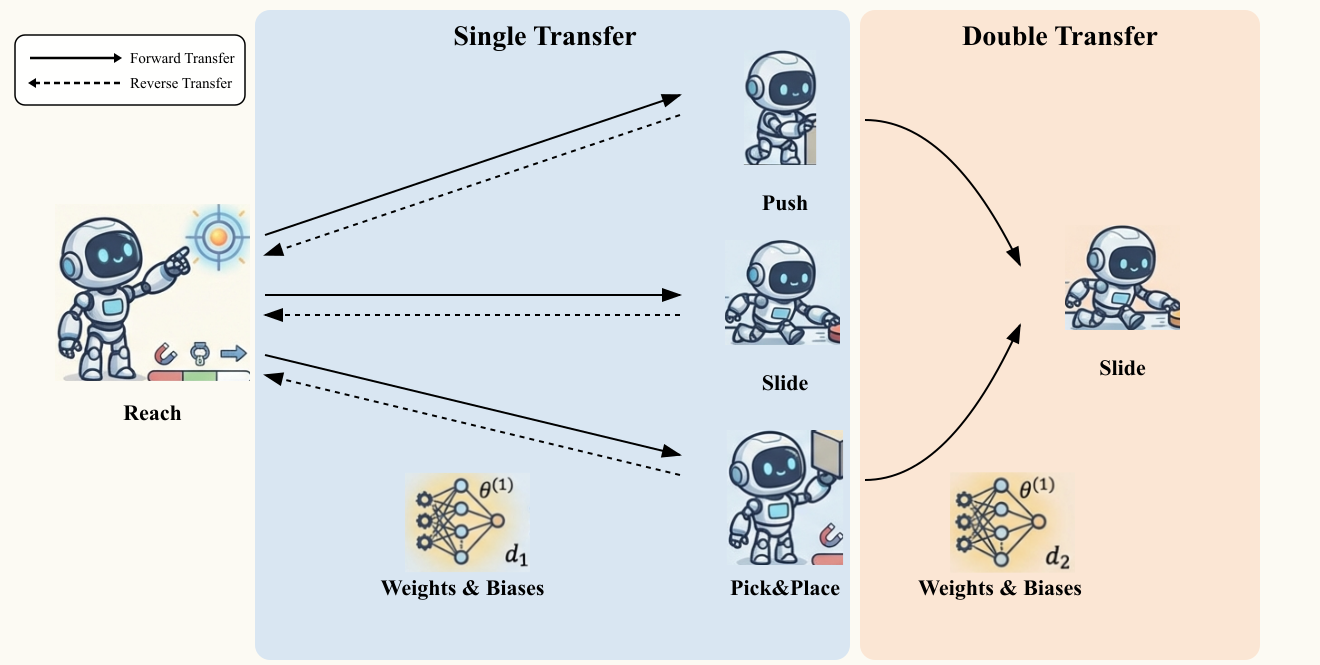}
    \caption{Transfer learning workflow with forward and reverse transfer across manipulation tasks in obstacle-free and obstacle environments. Weights and biases of the first layer of critic networks are transferred from source tasks to target tasks.}
    \label{TL_process}
    \vspace{-0.2cm}
\end{figure*}

\begin{algorithm}[t]
\caption{TQC with Task Decomposition (first-layer transfer + RR)}
\label{alg:tqc_tl_rr}
\begin{pcode}
\item \kw{Input:} source task $\tau_s$, target task $\tau_t$; TQC hyperparameters; optional pretrained critic $Q_{\psi,\tau_s}$; redesigned reward $\mathcal{R}_\theta$
\item \kw{Output:} trained goal-conditioned policy $\pi_\phi(a|s,g)$
\item Initialize actor $\pi_\phi$ and critic ensemble $\{Q_\psi^{(m)}\}_{m=1}^{M}$; initialize target critics $\{Q_{\bar{\psi}}^{(m)}\}_{m=1}^M$
\item \kw{if} a pretrained source critic $Q_{\psi,\tau_s}$ is available \kw{then}
\item \ind{1}\kw{for} $m=1$ \kw{to} $M$ \kw{do}\quad transfer first layer only: $\psi^{(m)}_{1,\tau_t} \leftarrow \psi^{(m)}_{1,\tau_s}$
\item Initialize replay buffer $\mathcal{D}$
\item \kw{for} iteration $=1$ \kw{to} $I$ \kw{do}
\item \ind{1}reset environment, sample goal $g$, obtain initial state $s$
\item \ind{1}\kw{for} environment step $=1$ \kw{to} $S$ \kw{do}
\item \ind{2}sample $a \sim \pi_\phi(\cdot|s,g)$ and execute; observe next state $s'$
\item \ind{2}$r \leftarrow \mathcal{R}_\theta(s,a,s',g)$; store $(s,a,r,s',g)$ in $\mathcal{D}$; set $s \leftarrow s'$
\item \ind{1}\kw{for} gradient step $=1$ \kw{to} $U$ \kw{do}
\item \ind{2}sample minibatch from $\mathcal{D}$; update critics (TQC loss) and actor (TQC objective); soft-update target critics
\item \kw{return} $\pi_\phi$
\end{pcode}
\end{algorithm}

Fig.~\ref{TL_process} illustrates the TL workflow. Algorithm~\ref{alg:tqc_tl_rr} summarizes the overall training process integrating first-layer transfer and RR within the TQC framework.

Let $\mathcal{T} = \{\tau_1, \tau_2, \dots, \tau_N\}$ denote a set of manipulation tasks that share identical state and action representations but differ in goal definitions and reward structures. Let $\tau_s \in \mathcal{T}$ denote a source task and $\tau_t \in \mathcal{T}$ a target task. Both tasks share the same state space $\mathcal{S}$ and action space $\mathcal{A}$ but differ in reward structure and goal configuration.

The first critic layer operates on the shared $(s,a,g)$ input, while deeper layers compose task-dependent value estimates. Shared input semantics motivate reusing the input layer, although they do not guarantee positive transfer when the obstacle geometry changes. Let $\psi^{(m)}_{l,\tau}$ denote layer $l$ of critic $m$ for task $\tau$. We copy
\begin{equation}
\psi^{(m)}_{1,\tau_t}\leftarrow\psi^{(m)}_{1,\tau_s},\qquad\forall m.
\end{equation}
All higher critic layers and the actor are randomly initialized. Every layer is optimized during target training, with no freezing or special fine-tuning schedule. The reported comparisons evaluate this initialization against training from scratch; they do not establish a complete ranking of transfer depths.

When selecting a source for a given target, we follow a simple-to-complex ordering over tasks that share the same $\mathcal{S}$ and $\mathcal{A}$ and exhibit compatible contact dynamics (e.g.\ Reach\arw{}Push, Push\arw{}Slide), preferring the source whose object-interaction pattern most closely matches the target. In addition to single transfer from simpler to more complex tasks, we also evaluate reverse and double transfer settings, following the same first-layer parameter transfer protocol. These transfer settings are defined as follows:

\emph{Forward Transfer} refers to transferring parameters from a simpler source task to a more complex target task.
This includes both single-step forward transfer (also referred to as \emph{Single Transfer}), where weights and biases of the first layer of critic networks are directly transferred from task $\tau_s$ to $\tau_t$ (e.g., Reach\arw{}Slide or Push\arw{}Slide), and multi-step forward transfer (also referred to as \emph{Double Transfer}), where parameters are transferred sequentially across multiple tasks (e.g., $\tau_1 \rightarrow \tau_2 \rightarrow \tau_3$), enabling progressive knowledge accumulation across tasks of increasing structural complexity. The objective is to reuse generalizable state--action representations learned under less constrained reward geometry to facilitate learning in more complex environments.

\emph{Reverse Transfer} performs the opposite direction (complex\arw{}simple), allowing us to examine whether representations learned under harder geometric and obstacle constraints can benefit simpler manipulation tasks.

\emph{Neutral Transfer} refers to the situation where transferring parameters from a source task does not lead to performance improvement on the target task compared to training from scratch. This may occur when discrepancies in the reward structure or task geometry cause the transferred representations to be insufficiently aligned with the target objective, particularly when task requirements differ (e.g., different object configurations).

\subsection{Decomposing Multi-step Tasks with Staged Reward Formulation}

To mitigate learning inefficiencies caused by basic dense reward signals, we introduce a staged reward formulation that decomposes the overall multi-step task objective into structured intermediate sub-goals, following evidence that staged reward shaping benefits long-horizon manipulation~\cite{Hundt2021}.

The redesigned reward function is defined as
\begin{equation}
R_{\theta}(s_t,a_t,g)
=
\sum_{i=1}^{K}
\lambda_i r_i(s_t,a_t,g),
\end{equation}
where each $r_i$ represents a sub-task reward and $\lambda_i$ is a scalar weighting coefficient.

In the Pick-and-Place task, the redesigned reward structure includes:
\begin{itemize}
    \item an \emph{approach reward}
    $r_{\text{app}} = -\|p_{ee} - p_{obj}\|_2$ encouraging the end-effector to approach the object,
    \item a \emph{grasp reward} $r_{\text{grasp}}$ activated upon successful contact and stable gripper closure,
    \item a \emph{transport reward} $r_{\text{move}}$ promoting progress toward the goal while maintaining collision-free motion.
\end{itemize}
The weights $\lambda_i$ are set manually to balance the magnitudes of the three terms and are kept fixed across all experiments. Specifying the reward requires identifying the ordered sub-goals and a per-term distance or contact signal, and this template transfers to other multi-step tasks (e.g., stacking or insertion) by re-listing their stages, whereas single-step tasks such as Push and Slide require no such decomposition, which is why RR is applied only to Pick-and-Place. The staged reward provides denser and more informative learning signals during early training and stabilizes convergence by guiding the policy through meaningful intermediate sub-goals.

\begin{table}[t]
\caption{Training hyperparameters for the DRL baselines and our framework. All obstacle and obstacle-free tasks share the same hyperparameters except for the total number of training steps.}
\label{tab:drl_hyperparams}
\centering
\footnotesize
\setlength{\tabcolsep}{4pt}
\resizebox{\columnwidth}{!}{%
\begin{tabular}{ll}
\toprule
\textbf{Category} & \textbf{Setting} \\
\midrule
Policy & MultiInputPolicy \\
Replay buffer & Off-policy replay buffer \\
HER strategy & Future (sparse-reward settings only) \\
Number of sampled goals & 4 (HER only) \\
Soft update coefficient $\tau$ & 0.05 \\
Replay buffer size & 1M \\
Batch size & 256 \\
Discount factor $\gamma$ & 0.95 \\
Learning rate & 0.001 \\
Network architecture & [512, 512, 512] \\
Number of critics $M$ (TQC) & 2 \\
\midrule
\multicolumn{2}{l}{\textbf{Total training steps (obstacle / obstacle-free)}} \\
\midrule
Push & 1M / 1M \\
Pick\&Place & 2M / 1M \\
Slide & 4M / 3M \\
Reach & 20K / 20K \\
\bottomrule
\end{tabular}}
\end{table}

\section{Experimental Results and Discussion}

\begin{table*}[t]
\caption{Success rate (mean$\pm$std over 3 runs) comparison with DRL baselines (DDPG, TD3, SAC, TQC) across all tasks under obstacle and obstacle-free settings. Bold marks the best mean per column.}
\label{tab:main_results}
\centering
\scriptsize
\setlength{\tabcolsep}{4pt}
\renewcommand{\arraystretch}{1.05}
\resizebox{0.82\textwidth}{!}{%
\begin{tabular}{lcccccc}
\toprule
& \multicolumn{2}{c}{Push} & \multicolumn{2}{c}{Pick\&Place} & \multicolumn{2}{c}{Slide} \\
\cmidrule(lr){2-3}\cmidrule(lr){4-5}\cmidrule(lr){6-7}
Algorithm
& Obstacle & No obstacle
& Obstacle & No obstacle
& Obstacle & No obstacle \\
\midrule
DDPG~\cite{lillicrap2020ddpg}
& $0.40\pm0.27$ & $0.93\pm0.10$
& $0.11\pm0.19$ & $0.16\pm0.06$
& $0.51\pm0.25$ & $0.64\pm0.03$ \\
TD3~\cite{fujimoto2018td3}
& $0.16\pm0.14$ & $1.00\pm0.01$
& $0.23\pm0.21$ & $0.20\pm0.04$
& $0.31\pm0.36$ & $0.37\pm0.32$ \\
SAC~\cite{Haarnoja2018sac}
& $0.43\pm0.04$ & $1.00\pm0.01$
& $0.33\pm0.08$ & $0.43\pm0.03$
& $0.84\pm0.06$ & $0.96\pm0.03$ \\
TQC~\cite{Kuznetsov2020tqc}
& $0.77\pm0.06$ & $\mathbf{1.00\pm0.00}$
& $\mathbf{0.67\pm0.22}$ & $0.98\pm0.02$
& $0.73\pm0.09$ & $0.93\pm0.04$ \\
\midrule
Ours
& $\mathbf{0.83\pm0.05}$ & $\mathbf{1.00\pm0.00}$
& $0.57\pm0.25$ & $\mathbf{0.99\pm0.01}$
& $\mathbf{0.95\pm0.02}$ & $\mathbf{0.98\pm0.01}$ \\
\bottomrule
\end{tabular}}
\end{table*}

\subsection{Simulation}

We evaluated the proposed task decomposition framework in the PyBullet Panda simulation environment on three manipulation tasks: \textit{Push}, \textit{Pick-and-Place}, and \textit{Slide}. Each task was implemented in two settings: obstacle-free and obstacle. In obstacle environments, a single cubic obstacle was introduced to increase task complexity. The obstacle center was sampled as
\[
x \sim \mathcal{U}(0,\,0.4), \quad
y \sim \mathcal{U}(-0.2,\,0.2), \quad
z = 0.05,
\]
with a minimum separation constraint of $0.1$\,m from both the object and the goal to avoid trivial configurations. The obstacle is a fixed-size cube whose position is randomized every episode, while its geometry and count are held fixed.

We compared against four off-policy DRL baselines: DDPG~\cite{lillicrap2020ddpg}, TD3~\cite{fujimoto2018td3}, SAC~\cite{Haarnoja2018sac}, and TQC~\cite{Kuznetsov2020tqc}. Unless otherwise specified, all DRL baselines shared identical hyperparameters, including a learning rate of $0.001$, batch size of $256$, discount factor $\gamma = 0.95$, soft update coefficient $\tau = 0.05$, and a replay buffer size of $1$M; the policy network was implemented as a three-layer MLP with $512$ units per layer, and $2$ critics were used for TQC.
For sparse-reward baselines, the HER mechanism adopted the `future' goal selection strategy with $4$ sampled goals per transition, which was employed to improve sample efficiency. HER is enabled for all sparse-reward settings and disabled whenever the dense redesigned reward (RR) is used, so that sparse and dense settings are never mixed within a single run.
Only the total number of training steps was adjusted according to task difficulty. Specifically, Push was trained for $1$M steps, Pick\&Place for $1$M (obstacle-free) and $2$M (obstacle) steps, Slide for $3$M (obstacle-free) and $4$M (obstacle) steps, and Reach for $20$K steps.
The shared hyperparameters were tuned for TQC and reused across algorithms. Reported convergence gains count target-task environment steps. End-to-end training also includes the source and, for double transfer, intermediate tasks: $S_{\mathrm{total}}=S_{\mathrm{source}}+S_{\mathrm{intermediate}}+S_{\mathrm{target}}$, with the intermediate term zero for single transfer. Reusing one source across $m$ targets amortizes its contribution to $S_{\mathrm{source}}/m$ per target. Thus, faster target learning and lower total training cost are distinct quantities.
Each experiment was repeated over $3$ independent runs with different random seeds. We report mean success rates and between-run standard deviations; the seed is the unit of replication.
All simulated experiments were run on a workstation equipped with an NVIDIA RTX 4060 GPU, and a supercomputer with an NVIDIA L40S GPU.
The training framework was based on Stable-Baselines3~\cite{Raffin2021} integrated with PyBullet. The complete hyperparameter configuration is summarized in Table~\ref{tab:drl_hyperparams}.

\begin{figure*}[t]
    \centering
    \includegraphics[width=0.30\textwidth]{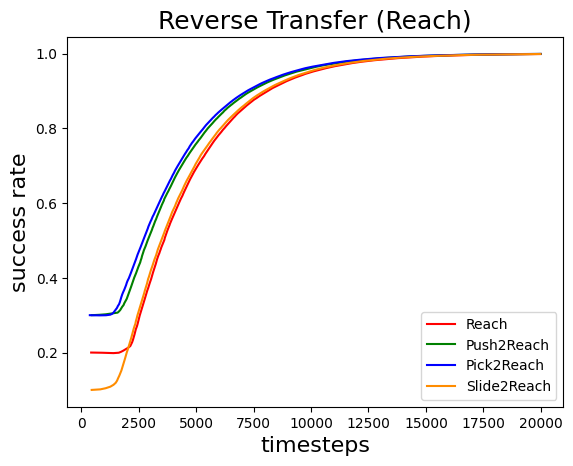}
    \includegraphics[width=0.30\textwidth]{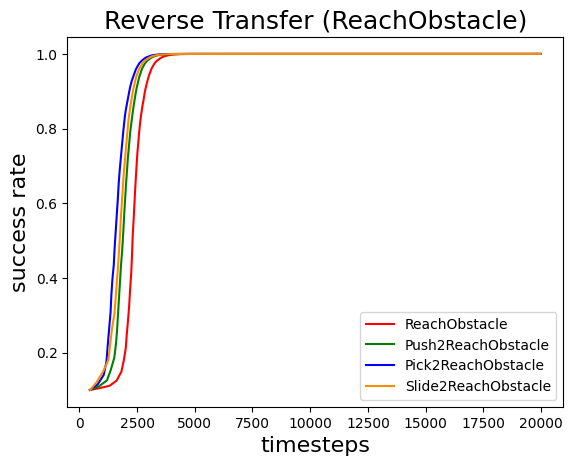}
    \\[2pt]
    \includegraphics[width=0.30\textwidth]{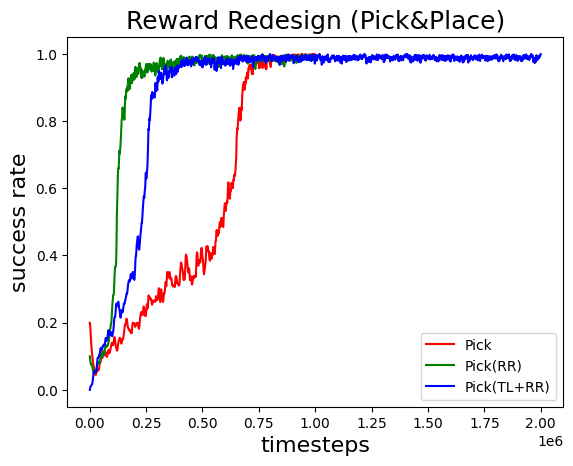}
    \includegraphics[width=0.30\textwidth]{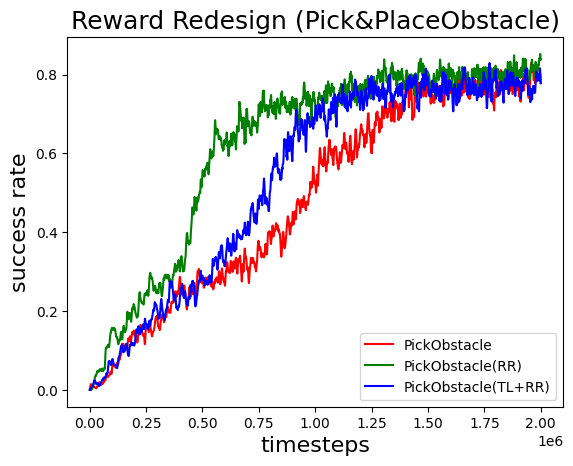}
    \caption{Reverse-transfer learning curves over three runs compared with the from-scratch TQC baseline for PandaReach (top row: obstacle-free, obstacle), and Reward Redesign results for PandaPick\&Place (bottom row: obstacle-free, obstacle).}
    \label{RT_RR}
    \vspace{-0.2cm}
\end{figure*}

Table~\ref{tab:main_results} summarizes the success rates over all baselines and our framework. Unless otherwise noted, \emph{Ours} reports the first-layer (forward) transfer configuration; the staged reward (RR), which applies only to the multi-step Pick-and-Place task, is reported and isolated separately in Fig.~\ref{RT_RR}.
In obstacle-free settings, most baselines achieve high performance on Push, indicating limited task complexity. However, performance gaps become evident in Pick\&Place and Slide. Our method achieves $0.99 \pm 0.01$ on Pick\&Place and $0.98 \pm 0.01$ on Slide, giving the highest reported mean in both settings.
The differences are more pronounced in obstacle environments. For Push with obstacle, the proposed method achieves $0.83 \pm 0.05$, exceeding all DRL baselines. On Slide with obstacle, our framework achieves $0.95 \pm 0.02$, substantially outperforming DDPG, TD3, and TQC, and improving over SAC. The one exception is Pick\&Place with obstacle, where performance variance increases across all methods and the transfer configuration ($0.57 \pm 0.25$) remains below TQC ($0.67 \pm 0.22$); we examine this case in detail below.

\begin{figure*}[t]
    \centering
    \includegraphics[width=0.30\textwidth]{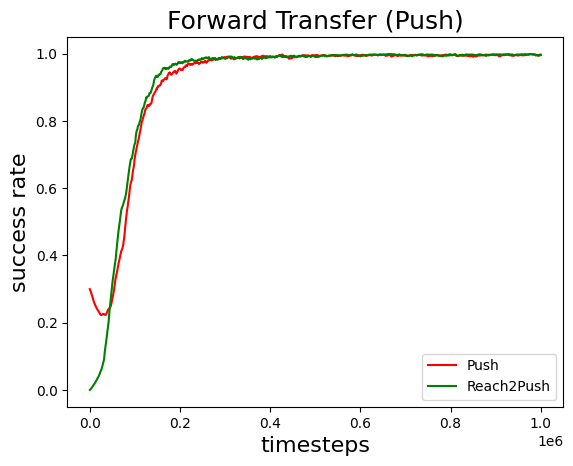}
    \includegraphics[width=0.30\textwidth]{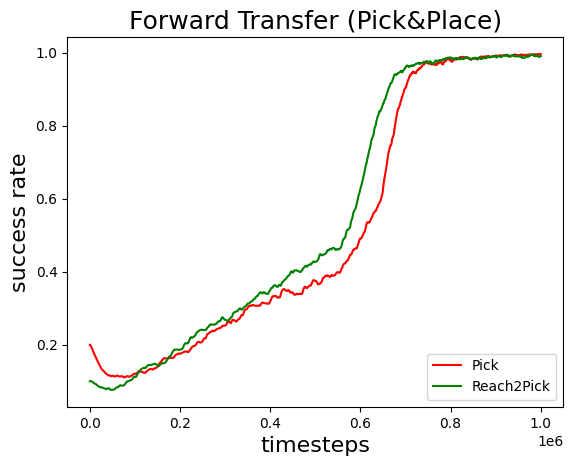}
    \includegraphics[width=0.30\textwidth]{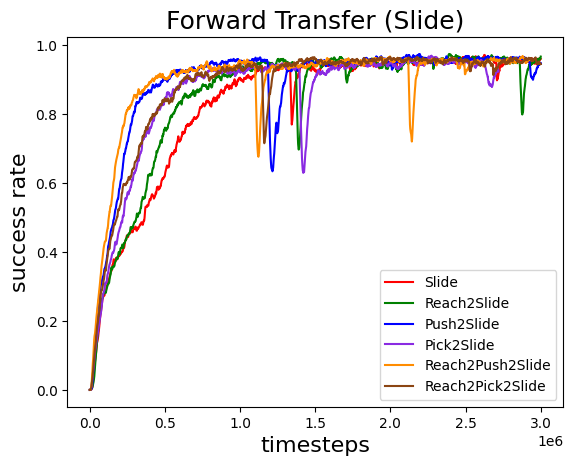}
    \includegraphics[width=0.30\textwidth]{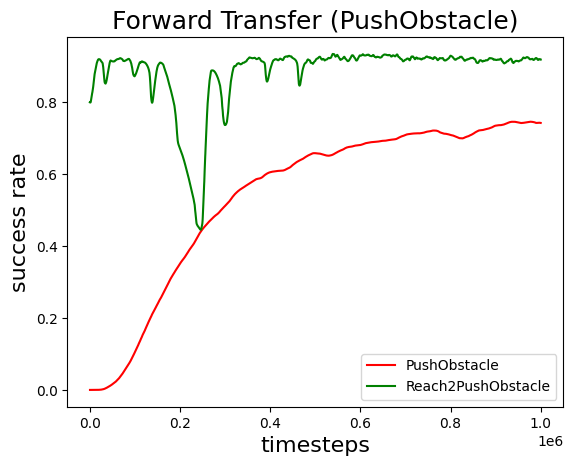}
    \includegraphics[width=0.30\textwidth]{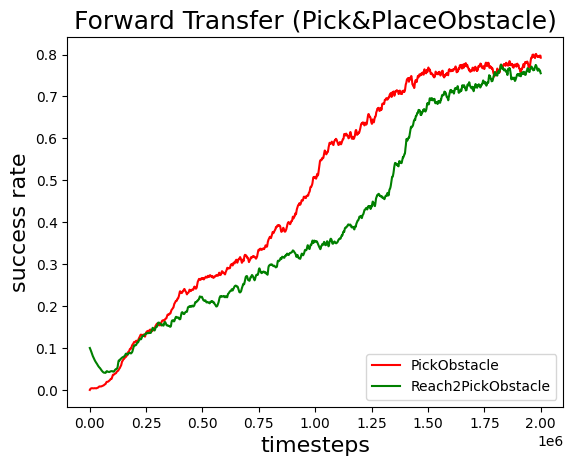}
    \includegraphics[width=0.30\textwidth]{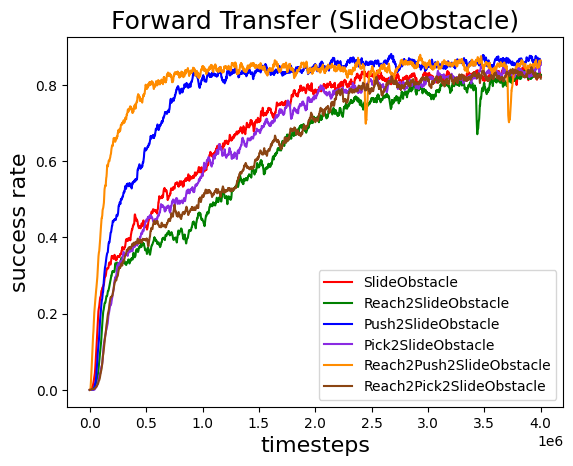}
    \caption{Forward-transfer learning curves over three runs compared with the from-scratch TQC baseline. Top row (obstacle-free): PandaPush, PandaPick\&Place, PandaSlide. Bottom row (obstacle): PandaPush, PandaPick\&Place, PandaSlide.}
    \label{FT}
    \vspace{-0.2cm}
\end{figure*}

Pick-and-Place with an obstacle couples grasping, transport, and collision avoidance. Transfer has a lower mean than TQC ($0.57\pm0.25$ versus $0.67\pm0.22$), with substantial between-seed variation. A possible explanation is that a reaching representation biases exploration toward approach directions made infeasible by the obstacle; the experiments do not isolate this mechanism. The RR comparison in Fig.~\ref{RT_RR} instead changes reward feedback on the target task. Its approximate curve plateau of $0.80$--$0.83$ is distinct from the mean-and-standard-deviation measurement in Table~\ref{tab:main_results}.

Table~\ref{tab:ablation} summarizes first-layer transfer versus training from scratch using task-specific success thresholds. The step ranges and plateaus are approximate readings of the learning curves in Figs.~\ref{FT}--\ref{RT_RR}, whereas Table~\ref{tab:main_results} reports success-rate means and standard deviations. These summaries describe different measurements and should be interpreted at their stated precision. The quantified transfer-depth evidence here concerns first-layer initialization versus scratch; it does not establish a complete ranking over all possible transfer depths.

\begin{table*}[t]
\caption{First-layer transfer vs.\ training from scratch: approximate training steps required to reach a task-specific success threshold (Thr.) and final training-curve plateaus, read from Figs.~\ref{FT}--\ref{RT_RR} over 3 runs.}
\label{tab:ablation}
\centering
\scriptsize
\setlength{\tabcolsep}{4pt}
\renewcommand{\arraystretch}{1.1}
\resizebox{0.82\textwidth}{!}{%
\begin{tabular}{llcccc}
\toprule
& & & \multicolumn{2}{c}{Steps to threshold} & Final plateau \\
\cmidrule(lr){4-5}
Target task & Transfer route & Thr. & Transfer & Scratch & Transfer / Scratch \\
\midrule
Push (obstacle) & Reach\arw{}PushObs & 0.7 & $\sim$0.1M & 0.5--0.7M & 0.88--0.90 / 0.70--0.75 \\
Slide (obstacle) & Reach\arw{}Push\arw{}SlideObs & 0.8 & 0.4--0.6M & 2.2--2.6M & 0.85--0.88 / 0.80--0.85 \\
Pick\&Place & Reach\arw{}Pick & 0.9 & 0.65--0.70M & 0.70--0.75M & $\sim$1.0 / $\sim$1.0 \\
Reach & Pick\arw{}Reach, Push\arw{}Reach & 0.8 & 4--5K & 6--7K & $\sim$1.0 / $\sim$1.0 \\
Reach (obstacle) & reverse transfer & 0.9 & 2.0--2.5K & 2.8--3.0K & $\sim$1.0 / $\sim$1.0 \\
Pick\&Place (obstacle) & Reach\arw{}PickObs & 0.7 & 1.5--1.6M & $\sim$1.3M & 0.76--0.78 / $\sim$0.80 \\
\bottomrule
\end{tabular}}
\end{table*}

\emph{Forward Transfer (Single vs Double).}
As detailed in Fig.~\ref{FT}, \textit{single transfer} improves convergence speed in a task- and constraint-dependent manner, with the most pronounced gains appearing in obstacle settings.
In PushObstacle, Reach\arw{}PushObstacle quickly attains a high-success regime, reaching $\sim0.7$ within the first $\sim10^5$ steps and holding a high plateau around $0.88$--$0.90$ for most of training, whereas training from scratch increases gradually, only approaching $\sim0.7$ after roughly $(5$--$7)\times10^5$ steps and plateauing lower at $\sim0.70$--$0.75$ by $1$M steps.
This indicates that transfer can simultaneously accelerate convergence and improve performance when obstacle-aware geometric priors from the source task are aligned with the target task.
In the less complex obstacle-free tasks, the benefit of single transfer is mainly a modest early-stage speed-up.
Pick\arw{}SlideObstacle consistently outperforms Reach\arw{}SlideObstacle, indicating that source tasks involving contact and object manipulation provide more compatible priors for downstream sliding dynamics than pure reaching behaviors.
\textit{Double transfer} becomes particularly beneficial because it provides a progressive structural bridge across tasks. In SlideObstacle, the two-step pipeline Reach\arw{}Push\arw{}SlideObstacle achieves the fastest and most stable improvement, reaching $\sim0.8$ within about $0.4$--$0.6$M steps and maintaining a plateau around $0.85$--$0.88$, whereas training from scratch reaches $\sim0.8$ much later (around $2.2$--$2.6$M steps) and the single-transfer Reach\arw{}SlideObstacle variant is slower still.
Although Reach\arw{}Pick\arw{}SlideObstacle also accelerates convergence relative to training from scratch, it remains inferior to Reach\arw{}Push\arw{}SlideObstacle in both convergence speed and final stability.
While Pick introduces grasp-related behaviors, Push more closely matches the planar object interaction and obstacle-aware contact patterns required in SlideObstacle, leading to stronger transfer efficiency.
In the obstacle-free Slide task, double transfer (e.g., Reach\arw{}Push\arw{}Slide) similarly enters the high-success regime earlier than training from scratch.

\emph{Reverse Transfer.}
Reverse transfer results (Fig.~\ref{RT_RR}, 1st, 2nd) show stable generalization from structurally complex tasks to simpler reaching behaviors.
For the obstacle-free Reach task (1st), Pick\arw{}Reach and Push\arw{}Reach reach $\sim0.8$ success around $4$--$5$K steps, while training from scratch reaches the same level at approximately $6$--$7$K steps, and all methods converge to near-perfect performance ($\sim 1.0$) by $12$--$15$K steps.
In ReachObstacle (2nd), transferred policies exceed $\sim0.9$ success within $2.0$--$2.5$K steps, compared to $2.8$--$3.0$K steps for the baseline, and all variants ultimately reach near $1.0$.
Since final performance remains unchanged, knowledge learned in more complex tasks transfers effectively to simpler reaching settings.

\emph{Unfavorable Transfer.}
In Pick\&PlaceObstacle (Table~\ref{tab:main_results}), Reach\arw{}PickObstacle exhibits a consistent lag behind training from scratch throughout most of its training time: the from-scratch policy reaches $\sim0.7$ success at around $1.3$M steps, whereas the transferred variant reaches $\sim0.7$ later (approximately $1.5$--$1.6$M steps), with the final plateau slightly favoring the baseline ($\sim0.80$ versus $0.76$--$0.78$).
This unfavorable-transfer behavior suggests that a reaching prior learned under simpler geometry may bias exploration toward suboptimal or collision-prone approach directions in obstacle-aware grasp-and-place sequences.

\emph{Reward Redesign.}
We further investigate the impact of RR and find that it markedly improves convergence speed in Pick-and-Place tasks, as shown in Fig.~\ref{RT_RR}. RR is primarily applied to multi-step settings such as Pick-and-Place, and is therefore not evaluated on single-step Push and Slide tasks. Here Pick(RR) corresponds to training from scratch with the shaped reward, which isolates the effect of RR from transfer.
In the obstacle-free setting (3rd), the effect is particularly pronounced: the sparse baseline requires approximately $0.70$--$0.75$M steps to exceed a success rate of $\sim0.9$, whereas Pick(RR) reaches $\sim0.9$ much earlier, around $0.18$--$0.22$M steps, and the combined TL+RR variant reaches $\sim0.9$ around $0.25$--$0.30$M steps, with all variants eventually saturating near $1.0$.
In the obstacle setting (4th), RR still provides a measurable acceleration but with reduced magnitude. PickObstacle(RR) reaches $\sim 0.7$ success at approximately $0.55$--$0.60$M steps, compared to around $1.0$M steps for the sparse baseline. The TL+RR variant reaches a similar level around $0.75$--$0.85$M steps. However, all curves converge to a comparable plateau near $0.80$--$0.83$ by the end of training.
We additionally observe that combining RR with TL does not consistently improve performance in obstacle settings. One possible explanation is that TL introduces suboptimal representations when task geometry differs significantly, leading to neutral transfer. In such cases, RR alone can provide more stable learning signals than RR+TL.
Unlike the obstacle-free case, reward shaping does not consistently increase the final success rate under geometric constraints, although improvements may be observed in individual runs.

\subsection{Real-world experiment}

\begin{figure*}[t]
    \centering
    \includegraphics[width=0.20\textwidth]{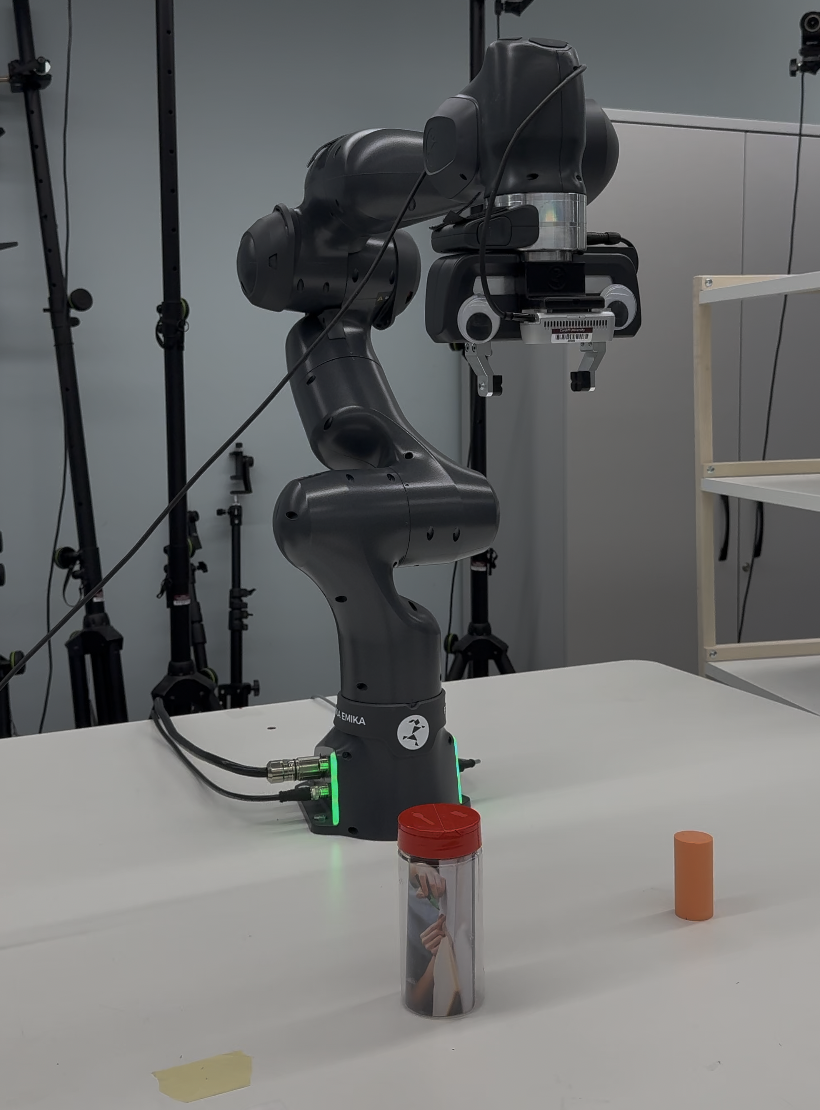}
    \includegraphics[width=0.20\textwidth]{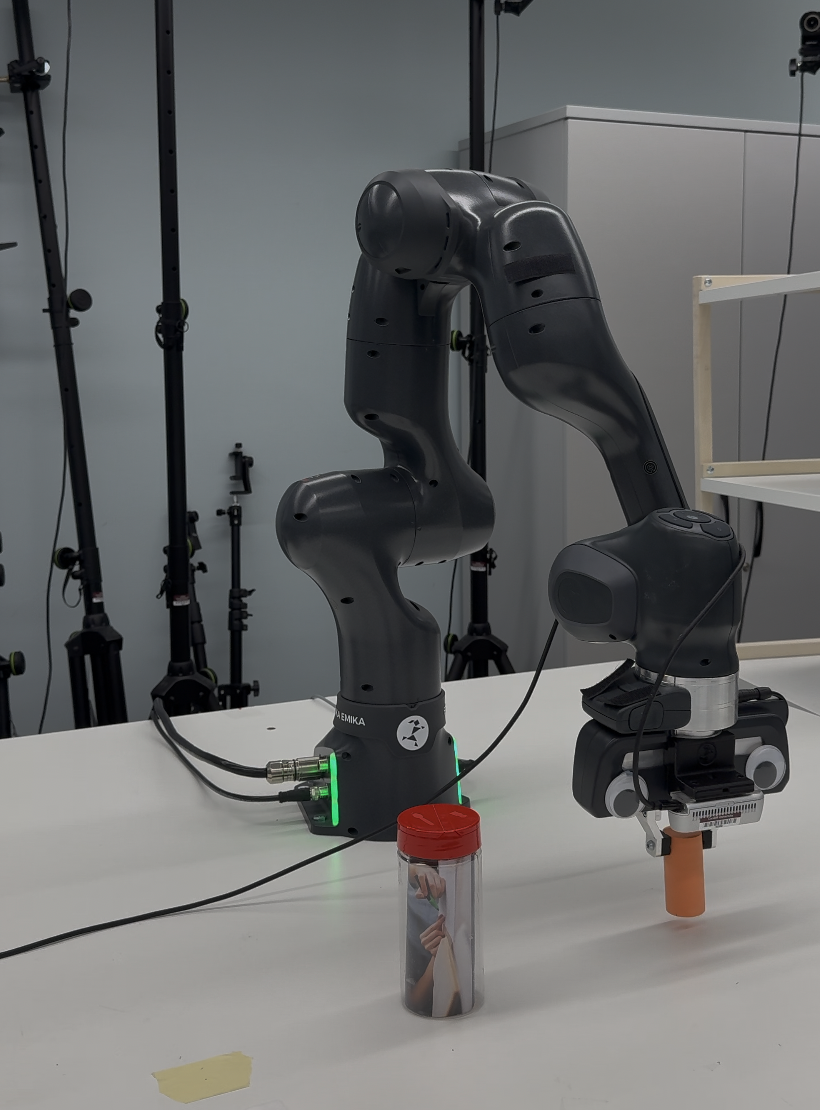}
    \includegraphics[width=0.20\textwidth]{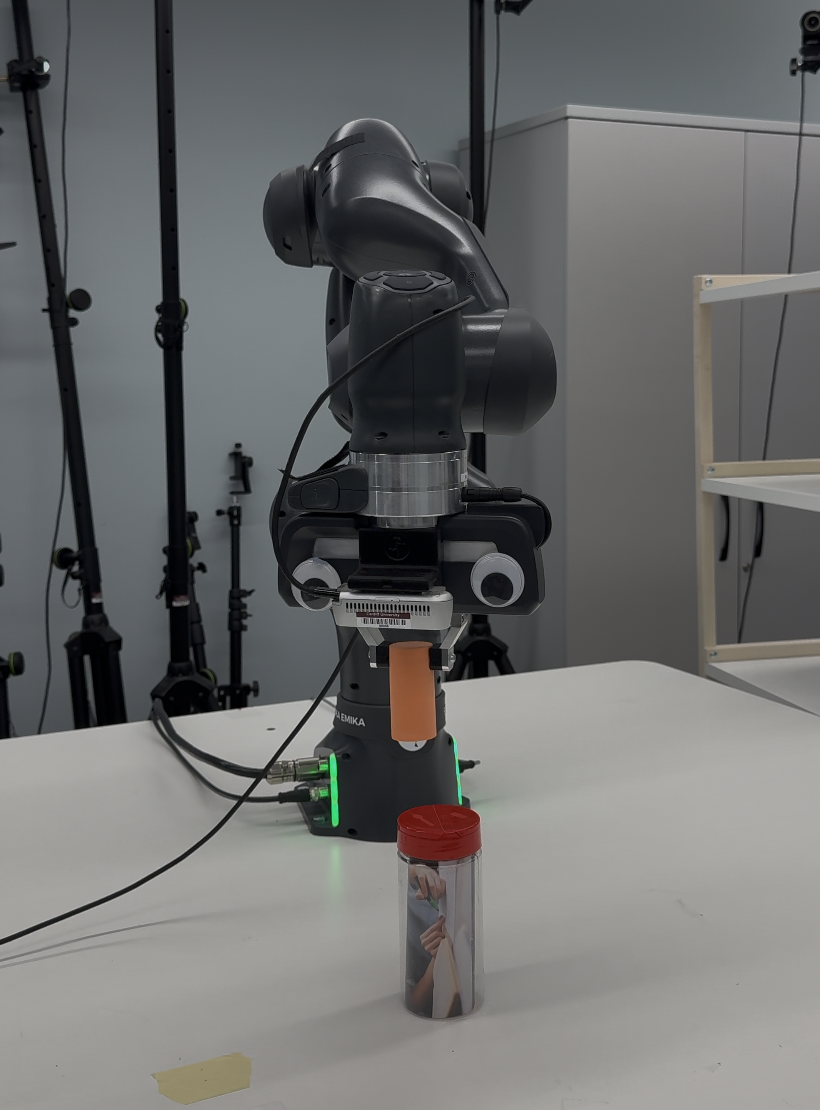}
    \includegraphics[width=0.20\textwidth]{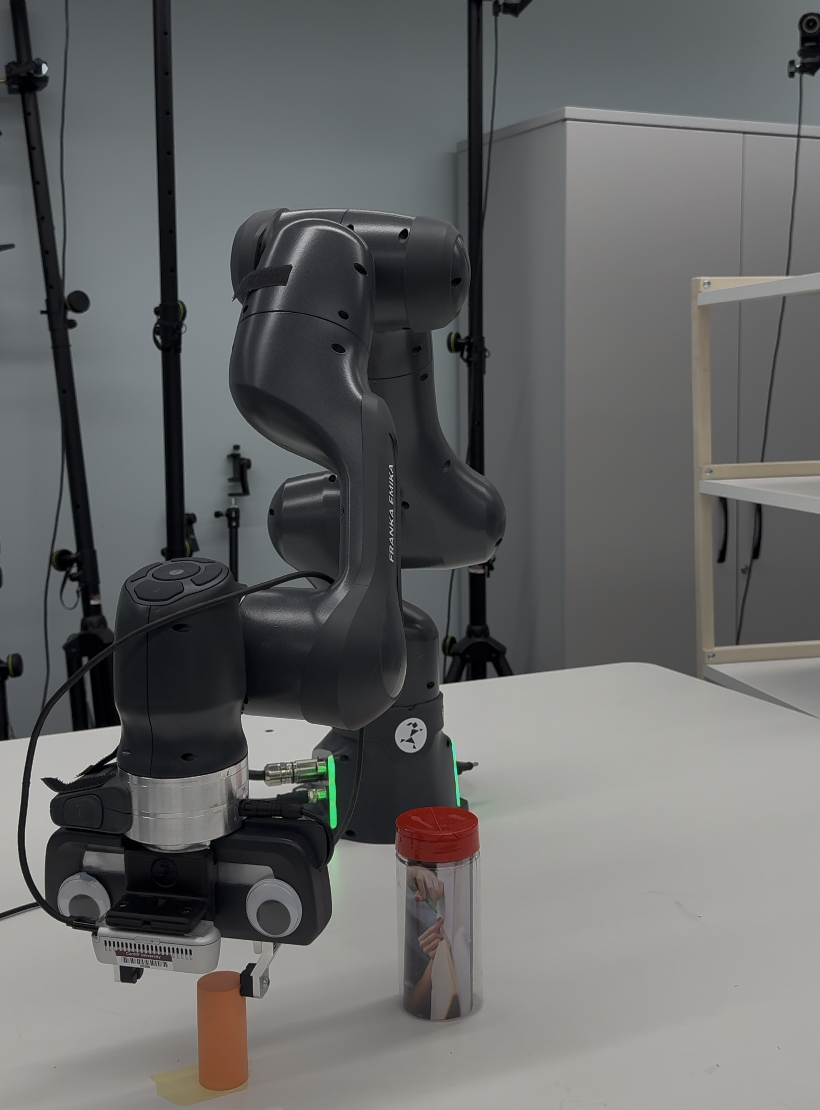}
    \caption{Real-world experiment on pick-and-place task with obstacle on Franka Emika FR3 robot.}
    \label{real_pp_obstacle}
    \vspace{-0.2cm}
\end{figure*}

We validate the proposed framework through a real-world experiment on a Franka Emika FR3 robot to assess its applicability to practical manipulation tasks. To ensure safe deployment and mitigate potential risks, a task-specific simulated environment is first constructed in PyBullet, which closely replicates the initial end-effector and object configurations of the real setup, including object geometries and spatial arrangements. A selected simulated trajectory is executed open-loop on the physical robot. This demonstrates trajectory feasibility; repeated closed-loop robot success rates are not evaluated.
Transferring trajectories from simulation to the real robot is challenging, particularly due to discrepancies in the kinematic structure and degrees of freedom between simulated and real systems. To alleviate this issue, a 7-DoF robotic arm model is employed in simulation to align joint-space representations with those of the FR3 robot. Furthermore, to address uncertainties arising from discretized trajectory intervals, linear interpolation is applied to segments with large joint displacements, ensuring smooth execution while satisfying the robot's kinematic and safety constraints.
An example of a real-world pick-and-place with obstacle task executed by the Franka arm is illustrated in Fig.~\ref{real_pp_obstacle}, and a video demo is available at our code link.

\section{Conclusion and Future Work}
First-layer critic transfer and staged RR improve learning in different parts of the tested task family. On obstacle-aware Push and Slide, transfer raises mean success by $6$ and $11$ percentage points relative to the strongest baselines. RR reduces target-training steps for obstacle-free Pick-and-Place from roughly $0.75$M to $0.20$M, about $73\%$ fewer steps. These results characterize a lightweight combination of established components on a TQC backbone.

Transfer benefits depend on the route and target constraints. An intermediate Push task provides a useful bridge to Slide, while reverse transfer accelerates reaching without changing its near-perfect final performance. In obstacle-aware Pick-and-Place, transfer alone gives a lower mean than training from scratch. RR accelerates early learning there, and adding transfer does not consistently improve on RR alone.

The three-seed experiments establish task-dependent learning behavior within the tested Panda environments. Source and intermediate training remain additional costs, and the Franka FR3 demonstration establishes feasibility of a selected open-loop trajectory. More seeds, complete transfer-depth comparisons, varied obstacle geometry, and repeated closed-loop robot trials would test how these findings extend to broader deployment conditions.

\balance
\bibliographystyle{IEEEtran}
\bibliography{References}

\end{document}